\documentclass{article}
\usepackage{spconf,amsmath,graphicx}
\usepackage{booktabs,multirow,makecell}
\usepackage{amsmath,amssymb,amsfonts}
\usepackage{subfig}

\usepackage{bibspacing}
\usepackage{enumitem}
\setlist{nosep, leftmargin=14pt}

\usepackage{mwe} % to get dummy images

\title{Spatial Temporal Graph Convolution with Graph Structure Self-learning for Early MCI Detection}
\name{Yunpeng Zhao$^1$, Fugen Zhou$^{1,2}$, Bin Guo$^{1\star}$, Bo Liu$^{1,2\star}$\thanks{$\star$Corresponding Authors. guobin420@gmail.com, bo.liu@buaa.edu.cn}}
\address{$^1$Image Processing Center, Beihang University, Beijing, China\\
$^2$Beijing Advanced Innovation Center for Biomedical Engineering, Beihang University, Beijing, China}

\begin{document}
%\ninept
%
\maketitle
\begin{abstract}
Graph neural networks (GNNs) have been successfully applied to early mild cognitive impairment (EMCI) detection, with the usage of elaborately designed features constructed from blood oxygen level-dependent (BOLD) time series. However, few works explored the feasibility of using BOLD signals directly as features. Meanwhile, existing GNN-based methods primarily rely on hand-crafted explicit brain topology as the adjacency matrix, which is not optimal and ignores the implicit topological organization of the brain. In this paper, we propose a spatial temporal graph convolutional network with a novel graph structure self-learning mechanism for EMCI detection. The proposed spatial temporal graph convolution block directly exploits BOLD time series as input features, which provides an interesting view for rsfMRI-based preclinical AD diagnosis. Moreover, our model can adaptively learn the optimal topological structure and refine edge weights with the graph structure self-learning mechanism. Results on the Alzheimer's Disease Neuroimaging Initiative (ADNI) database show that our method outperforms state-of-the-art approaches. Biomarkers consistent with previous studies can be extracted from the model, proving the reliable interpretability of our method.
\end{abstract}
\begin{keywords}
Alzheimer's Disease, Spatial Temporal Graph Convolution, Self-learned Graph Structure
\end{keywords}
\section{Introduction}
\label{sec:intro}

Alzheimer’s Disease (AD) is a cosmopolitan neurodegenerative disease in the brain and accounts for an estimated 60\%-80\% of dementia patients \cite{alzheimer20152015}. Progression of this irreversible and incurable disease is concomitant with gradually impaired cognitive skills, memory, and language abilities, leading to the escalation of potential social burden. Early mild cognitive impairment (EMCI) is a prodromal stage of AD with a high conversion rate \cite{li2021virtual}. Therefore, screening of EMCI is crucial for relieving the deterioration of AD. 

EMCI detection is still a challenging task. In light of trivial structural differences between EMCI and normal control (NC) brains \cite{kam2019deep}, recent studies seek resting-state functional MRI (rs-fMRI), a non-invasive imaging technique that measures blood oxygen level-dependent (BOLD) signals \cite{smitha2017resting}, to identify EMCI. According to related works, a traditional practice for EMCI screening with rs-fMRI is constructing features from BOLD signals and then applying machine learning methods to detect abnormal patterns. Commonly-used features include the whole-brain functional connectivity (FC) matrices, dynamic FC (dFC), and dynamic effective connectivity (dEC) extracted by group-constrained Kalman filter (gKF) algorithm \cite{li2021virtual,kam2019deep,bi2018random}. However, existing works overlooked the rich diagnostic information recorded in the spatial temporal dependency of original BOLD signals and have yet to explore the feasibility of building an end-to-end model that uses BOLD time series as features directly.

In fact, in rs-fMRI, spatially segregated brain regions are functionally connected with different weights and concurrently fluctuate along the temporal dimension, making it feasible to apply the graph neural networks (GNNs) to these data for classification \cite{kim2021learning, song2022multi}. For a GNN-based model, it is essential to construct a prior graph topological structure. Most existing methods tend to design a hand-crafted adjacency matrix that is consistent during training \cite{zhou2022interpretable, xing2019dynamic}. However, these approaches only focus on learning with the explicit prior brain structure, which may be nonoptimal and neglect implicit inter-region connections that contribute to EMCI detection. The ignorance of potential information from the implicit graph structure tends to yield incomplete brain structure modeling and limit the performance.

To overcome these limitations, we propose a novel spatial temporal graph convolutional network with graph structure self-learning mechanism for EMCI screening. Our contributions are three folds: 1) Our spatial temporal model can directly exploit BOLD time series as input features for EMCI detection by excavating spatial temporal dependencies in signals, which provides a new perspective for rsfMRI-based preclinical AD diagnosis. 2) To the best of our knowledge, our model is the first end-to-end GNN-based framework that can adaptively learn the optimal spatial dependency structure for EMCI detection. 3) The interpretability analysis of the self-learned graph topology could identify EMCI-contributory biomarkers consistent with existing neuroscience literature.

\section{Materials and Methodology}
\label{sec:method}

\subsection{Dataset and Preprocessing}
\label{ssec:data&preproc}

The neuroimaging data used in our study are obtained from the Alzheimer's Disease Neuroimaging Initiative (ADNI) database \cite{jack2008alzheimer}. Totally 146 age-gender-matched subjects are involved. All subjects have both rs-fMRI and T1 weighted images, which are acquired by 3 Tesla scanners from varying manufacturers, including Philips, Siemens, and General Electric (GE). Each subject has only one session. The demographic information of selected subjects is shown in Table.\ref{tab:demographicinfo}.
\vspace{-1em}
\begin{table}[h]
\caption{Demographic information of 146 subjects.}
\label{tab:demographicinfo}
\centering
\setlength{\tabcolsep}{7pt}
\begin{tabular}{*{3}{cc}}
\hline
\textbf{Group} & \textbf{NC} & \textbf{EMCI}\\
\hline
Gender (Male/Female) & 30/43 & 34/39\\
Age (Mean$\pm$SD) & 72.26$\pm$6.96 & 72.64$\pm$6.70\\
\hline
\end{tabular}
\end{table}

We utilize a widely-adopted DPARSF toolbox for image preprocessing. The procedures similar to \cite{kam2019deep} are as follows: 1) slice timing; 2) head motion correction; 3) coregister the T1 image to the functional image; 4) regress out nuisance covariates, including mean white matter and cerebrospinal fluid signals; 5) band-pass filtering ($0.01\leq f\leq 0.1\rm{Hz}$); 6) spatial normalization.

Afterward, we parcellate brain volumes into $N=116$ ROIs by the widely-used Automated Anatomical Labeling (AAL) template and then extract average time series from each ROI. As the neuroimages differ in volume numbers (140, 197, and 200 for Philips, Siemens, and GE scanners, respectively), we select the first 140 time points to equalize their size. All time series are transformed into \emph{z}-scores to remove amplitude effects. Finally, an ROI-wise average BOLD series in size of $116 \times 140$ is generated for each subject.

\begin{figure}[h]
\centering
\includegraphics[width=1.0\linewidth]{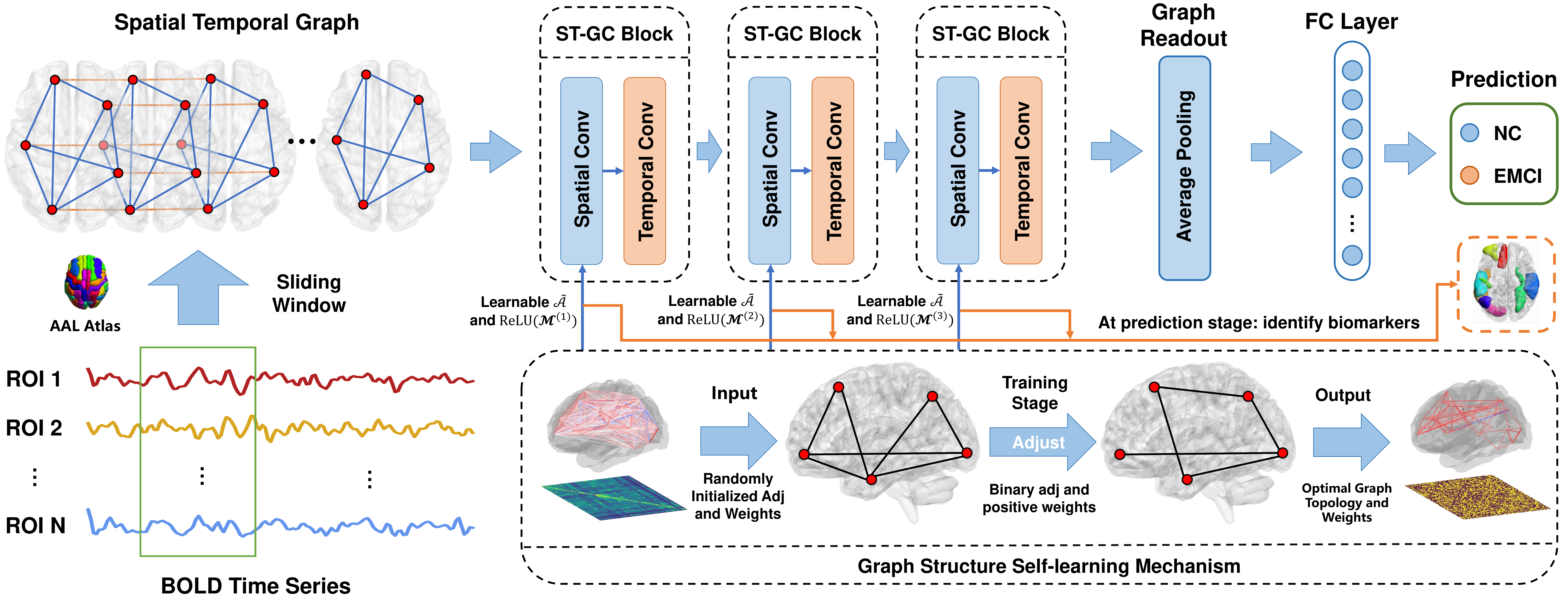}
\caption{The overall architecture of our proposed network.}
\label{fig:overview}
\end{figure}
\vspace{-1.2em}

\subsection{Spatial Temporal Graph Construction}
\label{ssec:graphconstruction}

For an ROI-wise average BOLD time series $f \in \mathbb R ^ {N \times Z}$ with \emph{N} ROIs and \emph{Z} time points, it can be naturally represented by an undirected spatial temporal graph $\mathcal G = (\mathcal V, \mathcal E)$ featuring both intra-brain connections and time series fluctuations. In the node set $\mathcal{V} = \left\{ v_{ti}|t = 1, ...,Z; i=1,...,N \right\}$, the $i$-th ROI at the $t$-th time point define a vertex $v_{ti}$, whose feature $\mathcal{F}(v_{ti})$ is the magnitude of the BOLD signal at $v_{ti}$. Therefore, the feature map of the spatial temporal graph is $\mathcal{F} \in \mathbb R ^ {C \times N \times Z}$, where $C=1$ denotes the dimension of node features. Meanwhile, the edge set $\mathcal E$ describes spatial and temporal connections between the nodes. For the spatial graph, we build edges for brain regions by randomly initializing the adjacency matrix, which is described in Sec.\ref{ssec:reweight}. The optimal spatial graph structure will be learned by the graph structure self-learning mechanism afterward. For the temporal graph, each ROI is linked to the same ROI in the next time point.

However, the spatial temporal dependencies are not stationary during the entire scan, and the pattern reveals considerable fluctuations. Inspired by sliding window-based dFC \cite{kam2019deep}, our model is trained on short sequences temporally sampled with window size $T$ to model such dynamics. Specifically, at each training iteration, we randomly sample a sub-sequence of length $T$ from the full BOLD time series of each subject in the mini-batch. At the testing stage, we apply the trained model to slices of each testing subject. We then average the sigmoid values to produce the final prediction.

\subsection{Spatial Temporal Feature Extraction}
\label{ssec:stgcn}
We define a spatial temporal graph convolution (ST-GC) block consisting of a graph convolution (GC) layer followed by a designed temporal inception module to generate discriminative spatial temporal representations of graph-structured BOLD signals. The GC layer aggregates spatial information at each time point according to the graph topology. Then the temporal inception module learns temporal features for each brain region independently. We concatenate several ST-GC blocks to extract spatial temporal features.

\textbf{Spatial Graph Convolution.} Noting that the spatial dependency is graph-structured, we employ the graph convolution operation in \cite{kipf2016semi} to fuse information between nodes and their neighbors at each time point. Given the adjacency matrix $\boldsymbol{A} \in \mathbb R ^ {N \times N}$ and the $H_1$-channel input feature map $\boldsymbol{X}_{1:N,1:T}^{(l)} \in \mathbb R ^ {H_1 \times N \times T}$, the spatial convolution at the $t$-th time point in the $l$-th ST-GC block is defined as
\begin{equation}\label{spatialconv}
\tilde{\boldsymbol{X}}_{1:N,t}^{(l)} = \boldsymbol{D}^{- \frac{1}{2}} \boldsymbol{A} \boldsymbol{D}^{- \frac{1}{2}} \boldsymbol{X}_{1:N,t}^{(l)} \boldsymbol{W}^{(l)},
\end{equation}
where $\tilde{\boldsymbol{X}}_{1:N,t}^{(l)} \in \mathbb R ^ {H_2 \times N}$ denotes the $H_2$-channel output feature map at time point $t$ after the spatial graph convolution, $\boldsymbol{D}$ is the degree matrix of $\boldsymbol{A}$, and $\boldsymbol{W}^{(l)} \in \mathbb R ^ {H_1 \times H_2}$ is a trainable weight matrix. The spatial graph convolution is performed at every independent time point, through which information of brain regions is spatially communicated.

\begin{figure}[htbp]
\centering
\includegraphics[width=0.7\linewidth]{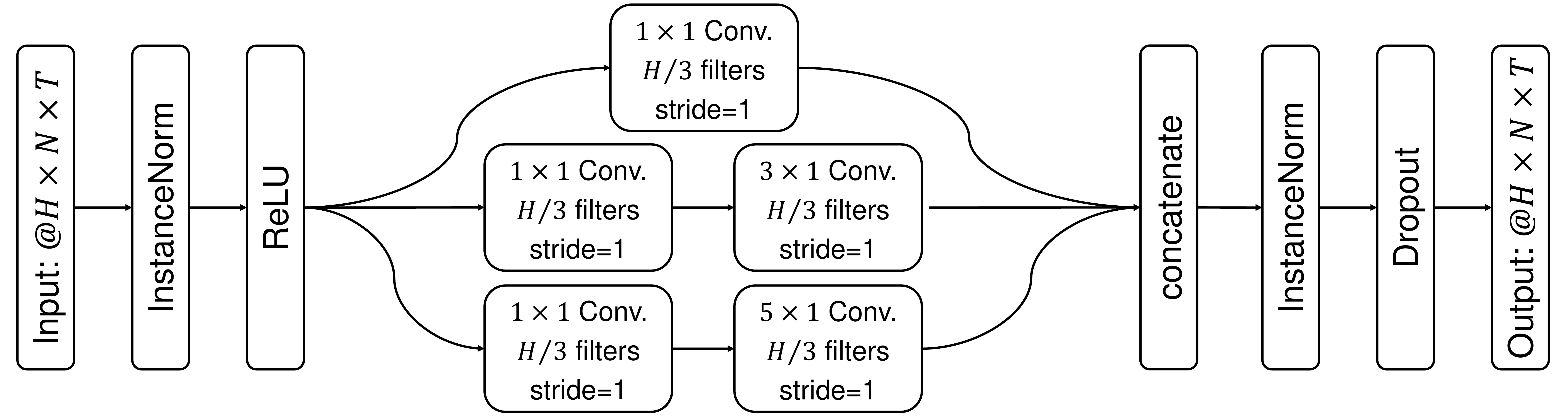}
\caption{Architecture of our temporal inception module.}
\label{fig:inception}
\end{figure}

\textbf{Temporal Inception Module.} Different from the spatial graph, the temporal graph possesses a grid structure. Therefore, it is feasible to adopt a standard 1D convolution for the temporal graph convolution. However, the kernel size could be too large or too small to excavate long-term and short-term patterns simultaneously. To further improve the temporal representation ability, we design a 1D temporal inception module inspired by \cite{szegedy2015going}. The architecture is shown in Fig.\ref{fig:inception}. We perform such temporal convolution on each brain region independently. Formally, let the features after spatial graph convolution in the $l$-th ST-GC block $\tilde{\boldsymbol{X}}_{1:N,1:T}^{(l)} \in \mathbb R ^ {H_2 \times N \times T}$ be the input feature map, our temporal graph convolution for the $i$-th ROI can be represented as
\begin{equation}\label{temporalconv}
\boldsymbol{X}_{i,1:T}^{(l+1)} = {\rm ReLU}({\rm TIM}_{\Omega} (\tilde{\boldsymbol{X}}_{i,1:T}^{(l)})),
\end{equation}
where $\boldsymbol{X}_{i,1:T}^{(l+1)} \in \mathbb R ^ {H_2 \times T}$ denotes the $H_2$-channel output of the $i$-th ROI after the temporal graph convolution, $\rm TIM$ is our temporal inception module with $\Omega$ representing its trainable parameters.

\vspace{-1.2em}
\subsection{Self-learned Graph Structure for EMCI Detection}
\label{ssec:reweight}
The graph structure self-learning mechanism learns the task-aware adjacency matrix adaptively to capture the optimal spatial dependencies and adjusts aggregation weights independently for each layer.

\textbf{Self-learned Graph Structure.} Existing GNN-based models for EMCI screening usually exploit a constant adjacency matrix and disregard the fact that the graph topology may be nonoptimal. Therefore, we design a dynamic adjacency matrix that is updated during training along with the model. Since the interactions between brain regions are mutual, the binary adjacency matrix is supposed to be symmetric. To reduce the computational complexity, we leverage a trainable vector $\boldsymbol{\Theta} \in \mathbb R ^ {N\times (N+1) / 2}$ to describe a lower triangular matrix, which is a compressed representation of a learnable symmetric matrix $\boldsymbol{\tilde{A}} \in \mathbb R ^ {N \times N}$. Specifically, we have $\boldsymbol{\tilde{A}}_{ij} = \boldsymbol{\tilde{A}}_{ji} = \boldsymbol{\Theta}_{i\times(i-1)/2+j}$, where $i,j \in [1,N]$. $\boldsymbol{\Theta}$ is randomly initialized before training. It can be easily seen that $\boldsymbol{\tilde{A}}$ is a complete graph that lacks interpretability and is computationally expensive. Therefore, we generate a sparse adjacency matrix $\boldsymbol{\mathcal{A}}$ by a soft-threshold operator
\begin{equation}\label{adjprob}
\boldsymbol{\mathcal{A}}_{ij} = {\rm ReLU}({\rm Sigmoid}(\boldsymbol{\tilde{A}}_{ij}) - {\rm Sigmoid}(\alpha)),
\end{equation}
where ${\rm Sigmoid}(\alpha) \in (0,1)$ is a threshold for sparsification. To eschew arbitrary thresholding, we also make $\alpha$ a trainable parameter initialized to 0. Noting that $\boldsymbol{\mathcal{A}}_{ij} \in \left[0,1\right)$, it can be considered as a probabilistic adjacency matrix, whose elements represent the probability to preserve the edge. Then after the sparsification, the problem is how to binarize $\boldsymbol{\mathcal{A}}$. The simple threshold approach is non-differentiable, thereby not feasible for stochastic gradient descent optimization. Therefore, we adopt the hard gumbel-softmax technique \cite{jang2017categorical} to conduct discrete sampling from $\boldsymbol{\mathcal{A}}$. In our paper, for the self-learning binary adjacency matrix $\boldsymbol{\tilde{\mathcal{A}}} \in \mathbb{R}^{N\times N}$, we have

\vspace{-0.4em}
\begin{footnotesize}
\begin{equation}\label{biadj}
\boldsymbol{\tilde{\mathcal{A}}}_{ij} = {\rm argmax}({\rm Softmax}(\frac{{\rm log}\boldsymbol{\mathcal{A}}_{ij}+g_{ij}^1}{\tau}, \frac{{\rm log}(1-\boldsymbol{\mathcal{A}}_{ij)}+g_{ij}^2}{\tau})),
\end{equation}
\end{footnotesize}%
where $g_{ij}^1,g_{ij}^2\sim {\rm Gumbel(0,1)}$, and $\tau$ is the temperature set to 0.2. At forward propagation, we use the binary adjacency matrix in Eq.\ref{biadj}. In contrast, at backward propagation, the $\rm argmax(\cdot)$ operation is abandoned, and gradients are calculated by softmax values. Finally, we successfully construct a self-learning adjacency matrix that is sparse and binary.

\textbf{Aggregation Weights Tuning.} We assume that connections between brain regions should contribute differently to EMCI screening. Therefore, we add a trainable weight matrix on every ST-GC block to scale the importance of spatial edges. Varying from the symmetric adjacency matrix, weight matrices are non-symmetric to model directional interactions between brain regions. Moreover, elements in the weight matrix should be non-negative because graph convolution based on message passing mechanism are incompatible with negative connection weights \cite{kan2021fbnetgen}. Hence, for the spatial graph convolution in the $l$-th ST-GC block, we rewrite Eq.\ref{spatialconv} to the form with the self-learned graph structure, which is
\begin{equation}\label{SLGS}
\tilde{\boldsymbol{X}}_{1:N,t}^{(l)} = \boldsymbol{D}^{- \frac{1}{2}} (\boldsymbol{\tilde{\mathcal{A}}} * {\rm ReLU}(\boldsymbol{\mathcal{M}}^{(l)})) \boldsymbol{D}^{- \frac{1}{2}} \boldsymbol{X}_{1:N,t}^{(l)} \boldsymbol{W}^{(l)},
\end{equation}
where $\boldsymbol{\mathcal{M}}^{(l)}$ is the learnable weight matrix in the $l$-th block which is initialized as all-ones, and $*$ denotes the element-wise product. Ultimately, Eq.\ref{SLGS} and Eq.\ref{temporalconv} jointly define a ST-GC block with the graph structure self-learning mechanism.

\vspace{-0.6em}
\subsection{Architecture and Optimization}
\label{ssec:archopt}
As shown in Fig.\ref{fig:overview}, we build a network consisting of $L=3$ ST-GC blocks to generate spatial temporal feature maps. Each ST-GC block has 64 channels for output, and the dropout rate is set to 0.5. Then the feature vector generated by a graph pooling layer is fed into a fully connected layer.

The method is developed using Pytorch on a single graphics card (i.e., NVIDIA RTX TITAN 12GB). We train the model using the Adam optimizer with a batch size of 16, a learning rate of 3e-4, and a weight decay of 1e-3. The window size $T$ is set to 12 time points (36s). For optimization, we adopt the binary cross-entropy loss $\mathcal{L}_{BCE}$. To highlight effective task-specific ROI connections, we encourage the sparsity of our self-learned graph structure by adding a sparsity regularization loss formulated as
\begin{equation}\label{sparloss}
\mathcal{L}_{SP} = \frac{1}{N\times(N+1)/2} \sum_{i=1}^{N\times(N+1)/2}{\rm Sigmoid}(\boldsymbol{\Theta}_{i}),
\end{equation}

Ultimately, our goal is to minimize the final loss function $\mathcal{L} = \mathcal{L}_{BCE} + \lambda\mathcal{L}_{SP}$, where $\lambda$ is a hyperparameter to adjust the weight of $\mathcal{L}_{SP}$ and is set to 1e-4.

\section{Experiments and Results}
\label{sec:expandres}

We perform experiments using the publicly available ADNI database. The detail of data acquisition is in Sec.\ref{ssec:data&preproc}. Stratified 10-fold cross validation is exploited to split the training set and test set. For hyperparameter tuning, we randomly select 10\% samples from the training set as the validation set in each fold. The performance of our model is measured by some common evaluation metrics: accuracy (ACC), area under the curve (AUC), sensitivity (SEN), and specificity (SPE). Results are reported by the mean plus/minus standard deviation across 10 test set splits.

\vspace{-0.5em}
\subsection{Performance Comparison with Relative Methods}
\label{cmpwithsota}

In this paper, we perform a binary classification task of NC vs. EMCI. In order to demonstrate the superior performance of our proposed method, we compare our model with some related state-of-the-art approaches with a similar number of samples. The classification results are shown in Table \ref{tab:cmpsota}. To ensure fairness, our comparison method is consistent with \cite{li2021virtual}. The comparison result demonstrates that our method achieves the best classification performance with 92.2\% ACC, 91.7\% SEN, and 92.9\% SPE, which are all significantly higher than other models. The AUC of our model also achieves 94.6\%.
\vspace{-0.5em}
\begin{table}[htbp]
  \centering
  \caption{Comparison with several state-of-the-art methods.}
  \label{tab:cmpsota}
  \resizebox{\linewidth}{!}{
  \begin{tabular}{*{6}{c}}
    \hline
    \textbf{Method} & \textbf{Subjects} & \textbf{ACC(\%)} & \textbf{AUC(\%)} & \textbf{SEN(\%)} & \textbf{SPE(\%)}\\
    \hline
    MK-SVM\cite{jie2018integration} & 50 NC, 56 EMCI & 78.3 & 77.1 & 82.1 & 74.0\\
    FSN-PFC\cite{yang2019fused} & 29 NC, 29 EMCI & 82.8 & 88.2 & - & -\\
	SF-net\cite{lei2020self} & 67 NC, 77 EMCI & 85.2 & 93.5 & 86.3 & 84.1\\
	SAC-GCN\cite{song2021graph} & 67 NC, 77 EMCI & 85.2 & 89.8 & 90.9 & 79.5\\
	MSGTN\cite{qiu2021multi} & 44 NC, 44 EMCI & 87.4 & 89.9 & 87.0 & 85.6\\
	cwGAT\cite{li2021virtual} & 72 NC, 53 EMCI & 90.9 & \textbf{96.7} & 90.4 & 91.4\\
	\textbf{Proposed} & 73 NC, 73 EMCI & \textbf{92.2} & 94.6 & \textbf{91.7} & \textbf{92.9}\\
    \hline
  \end{tabular}
  }
\end{table}
\vspace{-0.5em}

All the models compared require pre-determined brain networks or other features constructed from BOLD signals. In contrast, our method can learn the spatial temporal dependencies directly from BOLD time series and outperforms state-of-the-art methods.

\vspace{-0.5em}
\subsection{Ablation Study}
\label{ssec:ablation}

A set of ablation experiments are conducted to examine the effectiveness of different components. Specifically, we replace the self-learned graph structure with a fixed adjacency matrix in \cite{gadgil2020spatio}, replace our temporal inception module with a standard 1D convolution (kernel size=3, stride=1), remove the sparsity loss by set $\lambda=0$, and replace layer-wise weight matrices with a learnable matrix shared across all layers. Results can be seen in Table.\ref{tab:ablation}. The self-learned graph structure shows a significant impact ($\uparrow$ 2.1\% ACC) on the classification performance, since it allows the model to learn the effective task-aware spatial topological structure between brain regions. The sparsity loss also brings improvements ($\uparrow$ 0.6\% ACC) because it suppresses spurious edges and benefits node feature aggregation in spatial graph convolution. The temporal inception module ($\uparrow$ 0.7\% ACC) possesses various kernel sizes, which can model long-term and short-term temporal dependencies simultaneously in temporal graph convolution. Compared with a single learnable weight matrix that is shared across layers, layer-wise reweighting matrices improves the accuracy ($\uparrow$ 1.8\% ACC). This could be illustrated by the explanation that node representations in different layers are diverse, thereby requiring varying aggregation weights.

% \vspace{-1em}
\begin{table}[htbp]
\caption{Ablation study results.}
\label{tab:ablation}
\centering
\resizebox{\linewidth}{!}{
\begin{tabular}{{l}{c}{c}{c}{c}}
\hline
\textbf{Method} & \textbf{ACC(\%)} & \textbf{AUC(\%)} & \textbf{SEN(\%)} & \textbf{SPE(\%)}\\
\hline
\textbf{Proposed} & \textbf{92.2 $\boldsymbol{\pm}$ 2.3} & \textbf{94.6 $\boldsymbol{\pm}$ 2.5} & \textbf{91.7 $\boldsymbol{\pm}$ 3.1} & \textbf{92.9 $\boldsymbol{\pm}$ 2.9}\\
-w/o Self-learning Structure & 90.1 $\pm$ 1.9 & 92.0 $\pm$ 1.8 & 89.5 $\pm$ 2.7 & 90.3 $\pm$ 2.4\\
-w/o Temporal Inception & 91.6 $\pm$ 2.2 & 92.9 $\pm$ 2.5 & 90.2 $\pm$ 2.1 & 92.3 $\pm$ 2.0\\
-w/o Sparsity Loss & 91.5 $\pm$ 2.9 & 93.2 $\pm$ 2.8 & 90.3 $\pm$ 2.3 & 92.0 $\pm$ 2.7\\
-w/o Layer-wise Reweighting & 90.4 $\pm$ 3.0 & 91.8 $\pm$ 2.6 & 88.9 $\pm$ 2.9 & 91.7 $\pm$ 2.4\\
\hline
\end{tabular}}
\end{table}
\vspace{-1.5em}

\subsection{Interpretability}
\label{interpretation}

\begin{figure}[htb]
\centering
\begin{minipage}[b]{0.8\linewidth}
  \centering
  \subfloat[Adj Mat]{\includegraphics[width=0.24\textwidth]{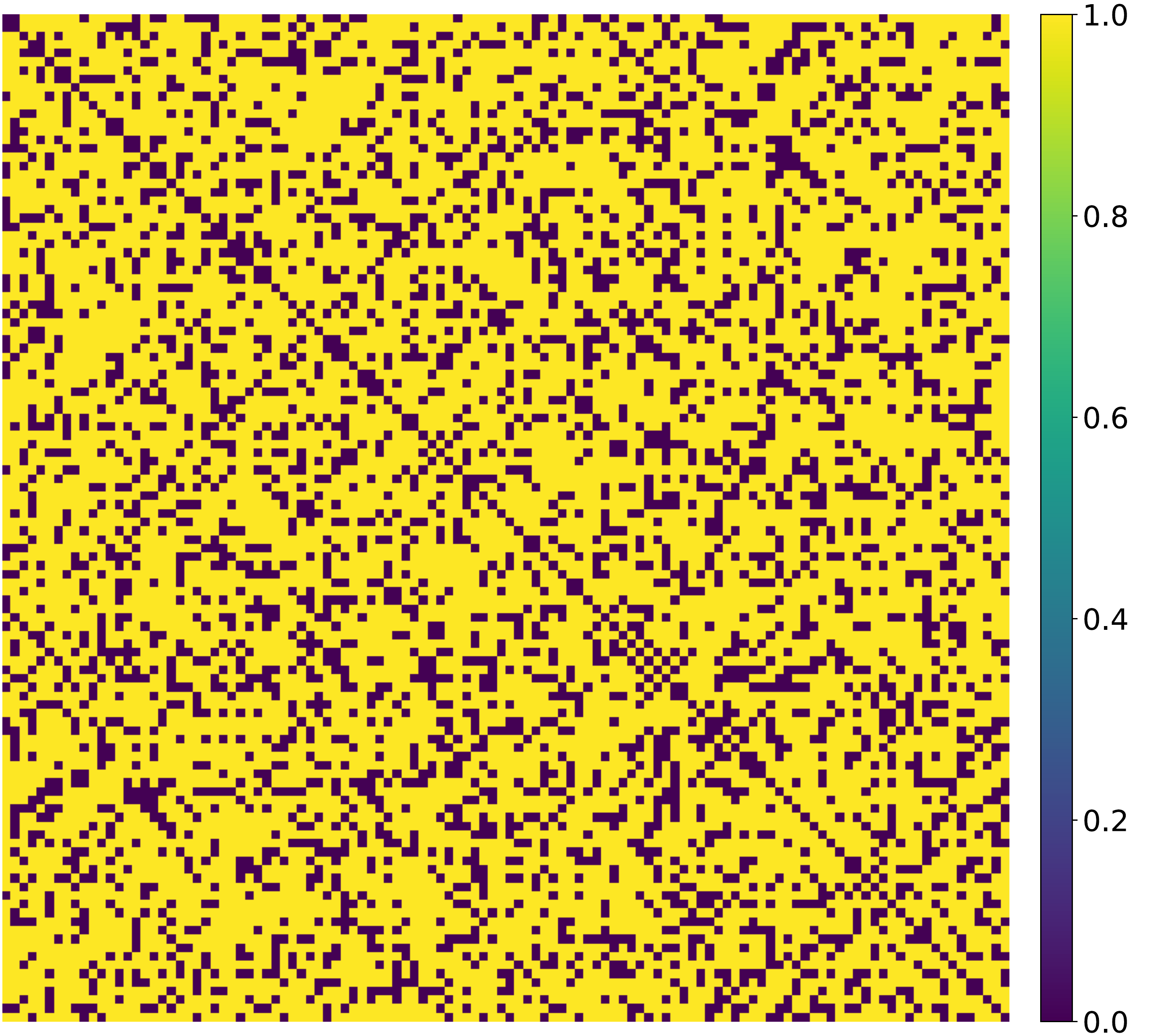}}
  \hfill
  \subfloat[Layer 1]{\includegraphics[width=0.24\textwidth]{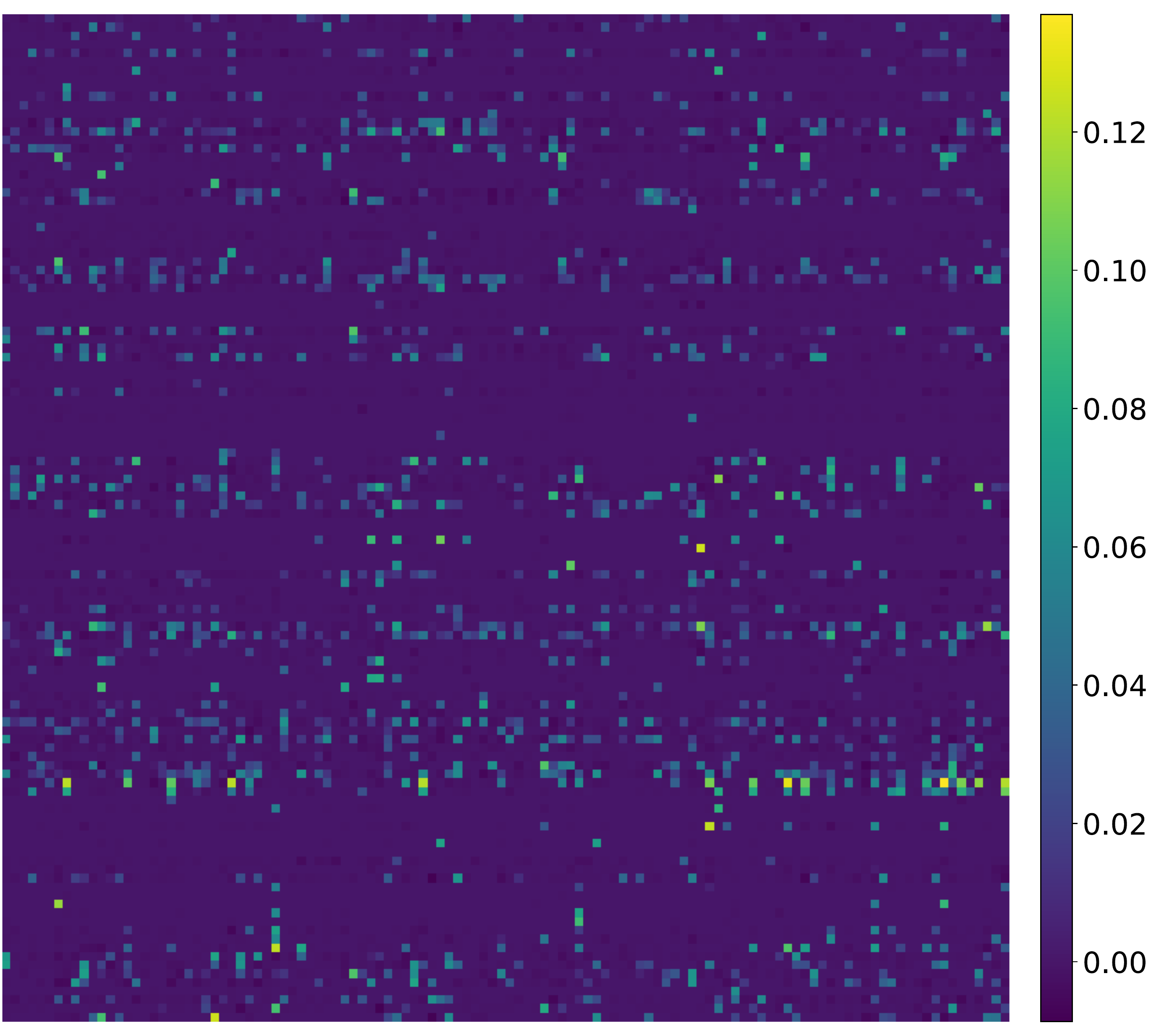}}
  \vspace{-1em}
  \subfloat[Layer 2]{\includegraphics[width=0.24\textwidth]{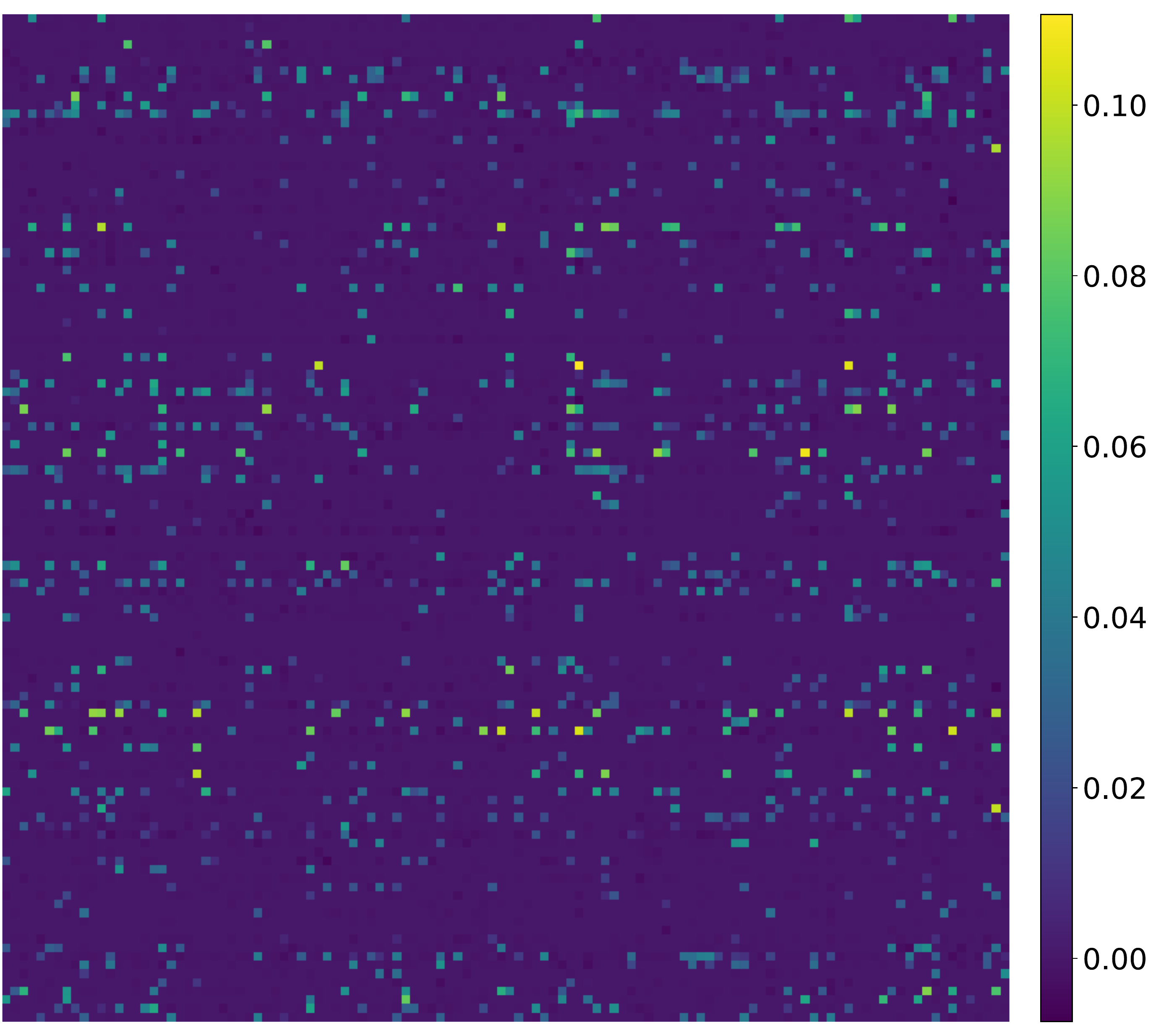}}
  \hfill
  \subfloat[Layer 3]{\includegraphics[width=0.24\textwidth]{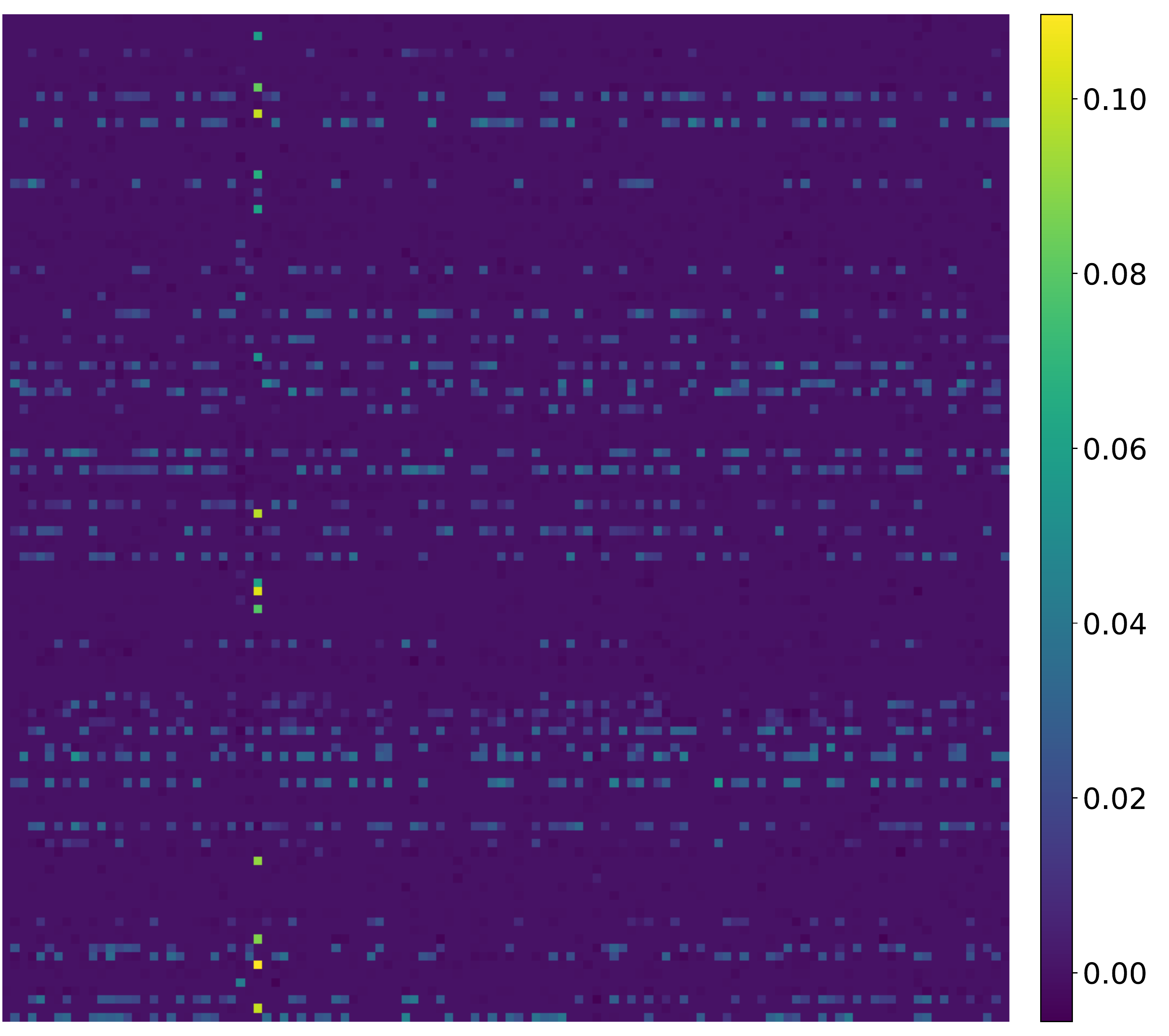}}
\end{minipage}
\caption{Learned adjacency matrix and aggregation weights.}
\label{fig:adjandweights}
\end{figure}

To figure out EMCI-contributory brain regions, we synthesize the learned graph structure and aggregation weights, shown in Fig.\ref{fig:adjandweights}. Specifically, we generate a score vector $\boldsymbol{\mathcal{S}} \in [0,1]^{N}$ for brain regions by $\boldsymbol{\mathcal{S}} = \sum_{i=1}^{N} \sum_{l=1}^{L} {\rm Norm}(\boldsymbol{\tilde{\mathcal{A}}} * {\rm ReLU}(\boldsymbol{\mathcal{M}}^{(l)}))$, where ${\rm Norm}(\cdot)$ represents Min-Max scaling.

\vspace{-1.5em}
\vspace{-0.5em}

Top 10\% salient ROIs are shown in Table.\ref{tab:salientrois}. The excavated ROIs, specifically including PUN.L, IFGoperc.R, ANG.R, MFG.R, ORBinf.L, MFG.L, and IFGtriang.L, are consistent with previous finding \cite{xiang2013abnormal}, proving the credible interpretability of our model. Besides, four ROIs from the cerebellum may suggest the possible involvement of the cerebellum in cognition \cite{jacobs2018cerebellum}.
\vspace{-0.5em}
\begin{table}[ht]
\caption{Top 10\% salient brain regions.}
\label{tab:salientrois}
\centering
\resizebox{\linewidth}{!}{
\begin{tabular}{{c}{c}{c}|{c}{c}{c}|{c}{c}{c}}
\hline
\textbf{ROI names} & \textbf{ROI index} & \textbf{score} & \textbf{ROI names} & \textbf{ROI index} & \textbf{score} & \textbf{ROI names} & \textbf{ROI index} & \textbf{score}\\
\hline
PCUN.L & 67 & 1.000 & Cerebelum.4.5.R & 98 & 0.828 & Cerebelum.10.L & 107 & 0.738\\ 
INS.R & 30 & 0.925 & MFG.R & 8 & 0.780 & ORBinf.L & 15 & 0.722\\
IFGoperc.R & 12 & 0.921 & Vermis.3 & 110 & 0.743 & MFG.L & 7 & 0.707\\
ANG.R & 66 & 0.838 & Vermis.9 & 115 & 0.741 & IFGtriang.R & 14 & 0.691\\
\hline
\end{tabular}}
\end{table}

\vspace{-1em}
\section{Conclusion}
\vspace{-0.5em}
\label{sec:conclusion}
This paper proposes a spatial temporal graph convolutional network with the graph structure self-learning mechanism, which directly utilize spatial temporal dependencies in BOLD time series to screen EMCI. Moreover, our graph structure self-learning mechanism successfully learns the optimal task-aware brain structure and aggregation weights. With these efforts, our model achieves better performance on the ADNI database compared with state-of-the-art methods. Further, we identify discriminative brain regions related to EMCI detection by analyzing learned graph structure and weights.

\section{Acknowledgments}
\label{sec:acknowledgments}
\vspace{-8pt}
This work was supported by the National Key R\&D Program of China under Grant Numbers:2018YFA0704100 and 2018YFA0704101, the National Natural Science Foundation of China (No. 61971443), and the Fundamental Research Funds for the Central Universities.

\vspace{-0.6em}
\section{Compliance with ethical standards}
\vspace{-8pt}
We wish to confirm that there are no known conflicts of interest associated with this publication. Ethical approval was not required, as confirmed by the license attached with the open access data.

\vspace{-0.6em}
\bibliographystyle{IEEEbib}
\bibliography{refs}

\end{document}